\documentclass[journal]{IEEEtran}
\usepackage{amssymb}
\usepackage{hyperref}
\usepackage{graphicx}
\usepackage[T1]{fontenc}
\usepackage{amsmath}
\usepackage{booktabs} 
\usepackage{multirow}
\usepackage{multicol}
\usepackage{color}

\ifCLASSINFOpdf
\else
\fi
\begin{document}
%
\title{DePF: A Novel Fusion Approach based on Decomposition Pooling for Infrared and Visible Images}
%
%
%
\author{Hui Li\textsuperscript{1*}, Yongbiao Xiao\textsuperscript{1}, Chunyang Cheng, Zhongwei Shen and Xiaoning Song
\thanks{This work was supported by the National Natural Science Foundation of China (62202205), the National Social Science Foundation of China(21$\&$ZD166), the Natural Science Foundation of Jiangsu Province, China(BK20221535), and the Fundamental Research Funds for the Central Universities (JUSRP123030).}
\thanks{Hui Li, Yongbiao Xiao, Chunyang Cheng and Xiaoning Song are with International Joint Laboratory on Artificial Intelligence of Jiangsu Province, School of Artificial Intelligence and Computer Science, Jiangnan University, 214122, Wuxi, China.}
\thanks{Zhongwei Shen is with School of Electronic and Information Engineering, Suzhou University of Science and Technology, 215009, Suzhou, China}
\thanks{1 Hui Li and Yongbiao Xiao contributed equally to this work and should be considered co-first authors.  \emph{(* Corresponding author: Hui Li. Email: lihui.cv@jiangnan.edu.cn.)}}
}
\markboth{Journal of \LaTeX\ Class Files,~Vol.~14, No.~8, June~2023}%
{Shell \MakeLowercase{\textit{et al.}}: Bare Demo of IEEEtran.cls for IEEE Journals}
%



\maketitle

\begin{abstract}
Infrared and visible image fusion aims to generate synthetic images simultaneously containing salient features and rich texture details, which can be used to boost downstream tasks.
However, existing fusion methods are suffering from the issues of texture loss and edge information deficiency, which result in suboptimal fusion results. Meanwhile, the straight-forward up-sampling operator can not well preserve the source information from multi-scale features.
To address these issues, a novel fusion network based on the decomposition pooling (de-pooling) manner is proposed, termed as DePF. Specifically, a de-pooling based encoder is designed to extract multi-scale image and detail features of source images at the same time. In addition, the spatial attention model is used to aggregate these salient features. After that, the fused features will be reconstructed by the decoder, in which the up-sampling operator is replaced by the de-pooling reversed operation. Different from the common max-pooling technique, image features after the de-pooling layer can retain abundant details information, which is benefit to the fusion process. In this case, rich texture information and multi-scale information are maintained during the reconstruction phase. The experimental results demonstrate that the proposed method exhibits superior fusion performance over the state-of-the-arts on multiple image fusion benchmarks.

\end{abstract}

\begin{IEEEkeywords}
image fusion, decomposition pooling, multi-scale features, detail features, deep learning.
\end{IEEEkeywords}

%
\IEEEpeerreviewmaketitle

\section{Introduction}

\IEEEPARstart{I}{mage} fusion task aims to extract complementary information from source images and generate composite images containing rich information. As an important image processing technology, image fusion has been widely used in digital remote sensing~\cite{li2018fusing}, classification~\cite{gao2018object}, object detection~\cite{guo2021deep,gao2021unified} and object tracking~\cite{zhang2019multi,zhu2023visual}. For infrared and visible image fusion task, the infrared modality contains salient thermal radiation information but fewer details, while the visible modality involves rich texture information. Therefore, how to preserve these complementary information from these two modalities is the key issue. 

Before the rise of deep learning, most image fusion methods are based on signal processing operations, such as multi-scale transform (MST)~\cite{nencini2007remote,lewis2007pixel} based methods, sparse representation (SR)~\cite{pati1993orthogonal,li2012group,chen2014image,yang2012pixel} and low-rank representation (LRR)~\cite{li2018infrared,gao2021improving,bhavana2021infrared} based methods. Non-deep learning methods usually include the following three steps: feature extraction, feature fusion and image reconstruction. Although these image fusion methods can synthesize satisfactory images in certain scenarios, they still contain limitation. On the one hand, manually designed fusion strategies cannot adapt to complex scenes. On the other hand, when extracting features from multi-modal images, the feature differences of each modality are usually not considered, which is difficult to comprehensively capture the features of source images. These problems prevent non-deep learning methods from being extended in practical applications.\par

With the development of deep learning, many deep learning based fusion methods are proposed to address the above drawbacks and the fusion performance has been improved. Deep learning based image fusion methods can be roughly divided into two categories: auto-encoder based methods\cite{li2018densefuse,li2020nestfuse} and end-to-end fusion networks\cite{ma2019fusiongan,li2023lrrnet}. 

\begin{figure}[!htbp]
\centering
\includegraphics[width=0.4\textwidth]{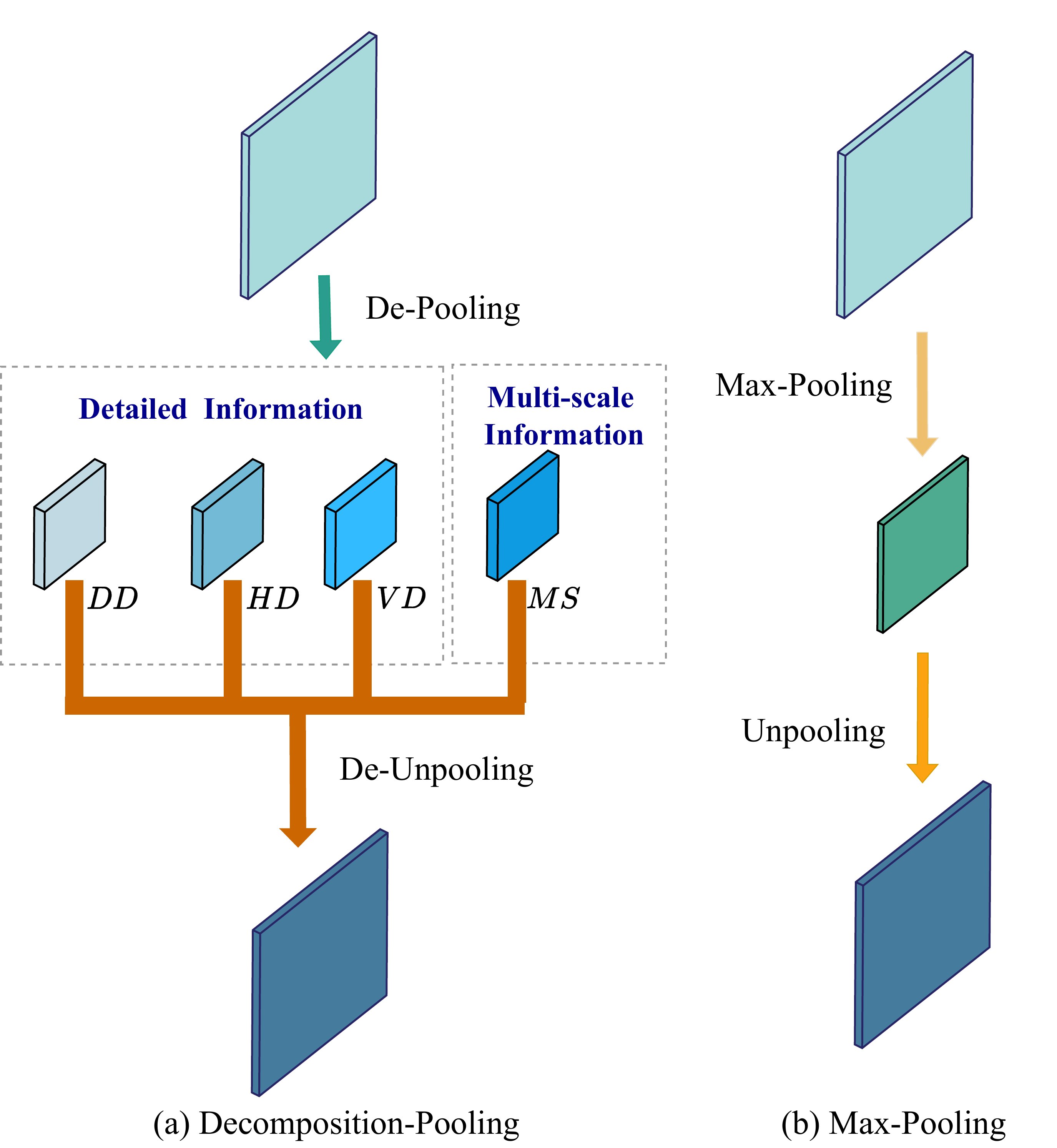}
\vspace{-0.2cm}
\caption{Comparison between de-pooling based architecture (a) and max-pooling based model (b). Given the features, our de-pooling can preserve multi-scale features and detail features simultaneously, but the max-pooling operation can only extract multi-scale feature information}\label{framework-max-wave}
\vspace{-0.2cm}
\end{figure}

In recent years, some typical auto-encoder based networks have been proposed~\cite{li2021rfn,li2023aefusion}. The auto-encoder architecture is introduced into image fusion mainly because of the insufficient multi-modal dataset, which is difficult to train a complex end-to-end fusion network. With the training strategy of auto-encoder, a large of single modality datasets can be used to train the encoder (feature extraction) and decoder (image reconstruction)\cite{li2018densefuse}. Then, manually designed fusion strategies are applied into auto-encoder framework in testing phase. The auto-encoder based fusion methods are generally composed of an encoder, a fusion strategy block and a decoder. These methods can make full use of the excellent nonlinear fitting ability of neural networks and utilize unsupervised ways to improve the quality of generated images. 
However, most of them are unable to sufficiently extract features, and the manually designed complex fusion strategies are only suitable for specific scenarios.\par

With the increase of multi-modal datasets, it is possible to design the end-to-end fusion network and the learnable fusion strategies. The generative adversarial network (GAN)~\cite{2014Generative}based methods are ideal for unsupervised fusion task. FusionGAN~\cite{ma2019fusiongan} is the first method to apply GAN into image fusion tasks. 
In FusionGAN, the generator aims to generate fused images with salient features and visible gradients, and the discriminator aims to force the fused image to preserve more texture details from visible image. However, the training of GAN networks is unstable and this method does not focus on downstream tasks. Therefore, in order to fit the high-level vision tasks, such as semantic segmentation and object detection, SeAFusion~\cite{tang2022image} cascades the image fusion module and semantic segmentation module, and simultaneously designs a gradient residual dense block to enhance fine-grained spatial details. Furthermore, more and more networks focus on the fusion of nighttime images. DIVFusion~\cite{tang2023divfusion} designs two modules to remove the illumination degradation and enhance the contrast and texture details of the fused features respectively. It reasonably lights up the darkness and facilitates complementary information aggregation. Moreover, most methods only focus on the features from the source images and ignore the intermediate output of the network itself. MUFusion~\cite{cheng2023mufusion} introduces a novel memory unit architecture that mainly utilizes intermediate fusion results obtained during training to further supervise the fused images.

Although current end-to-end fusion methods can produce considerable results in most scenarios, there are still some drawbacks. Firstly, the lack of scale transformation in most fusion networks will lead to the inability to change the spatial scale of deep features. Secondly, some methods that introduce pooling to enrich features by scale transformation will bring a lot of information loss and artifacts. Finally, this kind of networks are not easy to train and the results are relatively unstable.



\begin{figure}[!htbp]
\begin{center}
\includegraphics[width=0.75\textwidth]{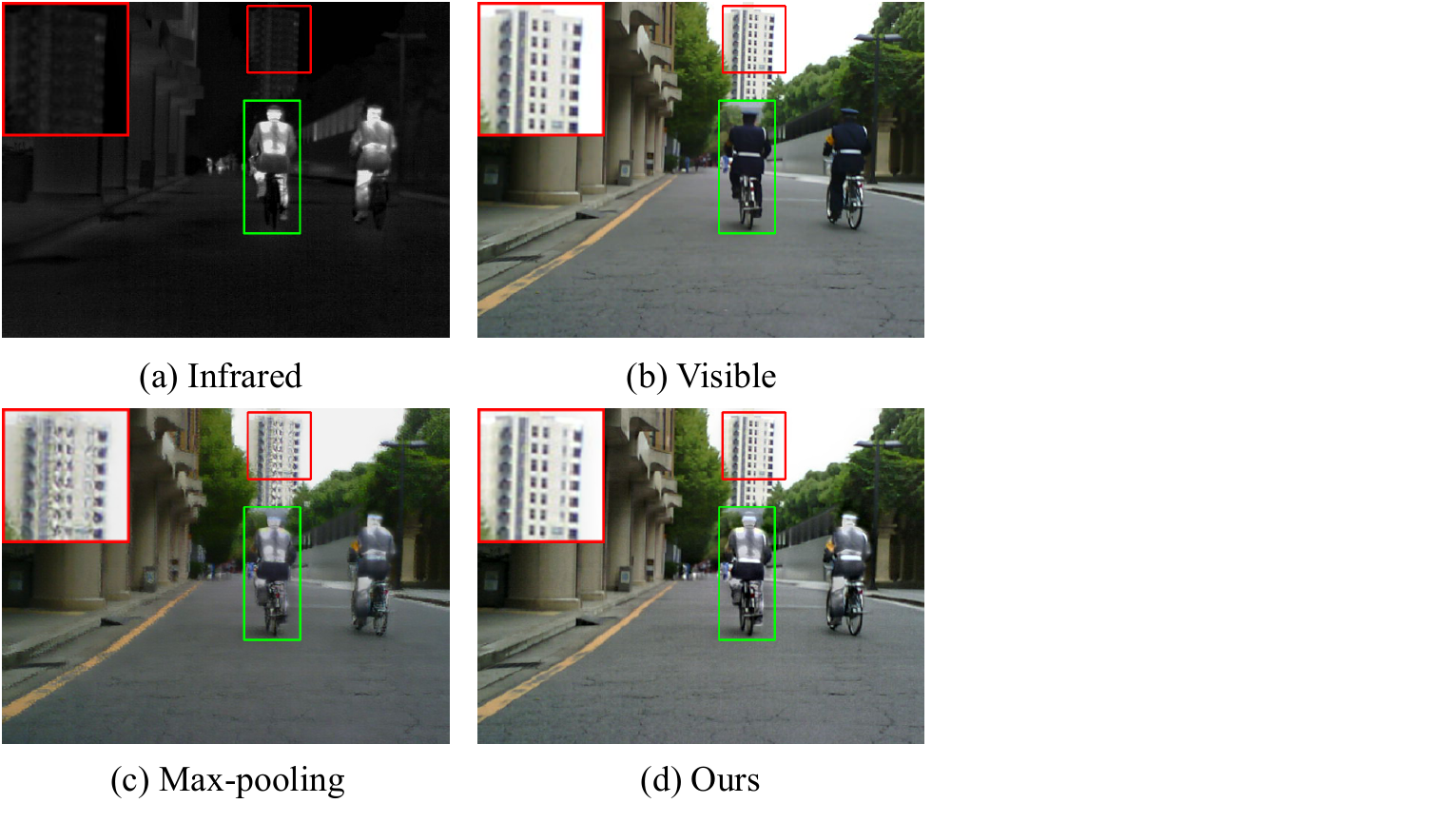}
\end{center}
\begin{center}
\vspace{-0.5cm}
\caption{Comparison between visualization results of the max-pooling based fusion method and the de-pooling (ours) based fusion method.}\label{compare-max_wave}
\end{center}
\end{figure}

To extract the multi-scale deep features, the existing fusion networks only utilize the pooling operation, such as max-pooling (Figure~\ref{framework-max-wave} (b)). Although it can extract multi-scale feature information, a lot of features are omitted which are important for fused image reconstruction. To address the above issue, we propose a novel decomposition pooling (de-pooling, Figure~\ref{framework-max-wave} (a)). Compared with max-pooling, de-pooling can extract multi-scale information without information omits. In addition, as shown in visualization results in Figure~\ref{compare-max_wave}, although most methods are valid by using max-pooling operations, they are not able to solve the information loss and artifacts. On the contrary, our method (Figure~\ref{compare-max_wave} (d)) avoids these drawbacks while enhancing the texture details.

The previous wavelet based convolutional networks~\cite{oyallon2017scaling,fujieda2018wavelet,williams2018wavelet,wang2021ldw} do not change the form of the kernel and they are mostly based on the kernel of the traditional wavelet transform. Compared with it, our method improves each kernel $\left( MS, VD, HD, DD\right)$ based on the Prewitt operator (details in \ref{depooling}). In addition, the receptive field of these methods is mostly $2\times2$, but the receptive field of the $4\times4$ we use is larger, which can maintain fine-grained features (details in \ref{fields}).

In our fusion framework, the proposed de-pooling is applied into encoder to extract multi-scale features and detail information at each scale. For the fusion strategy, the spatial attention model~\cite{li2020nestfuse} is introduced to fuse complementary information and finally the decoder is performed with multi-scale information and detail textures of the source images to reconstruct features. In the decoder, inverse de-pooling (de-unpooling) is utilized to inject the fused detail information while up-sampling the deep features. 
Due to its property of minimal information loss, the de-unpooling is able to completely reconstruct images without any post-processing step, which is crucial for image generation. The major contributions of this study are summarized as follows:
\begin{itemize}
\item A novel pooling operation named de-pooling is designed. It can extract multi-scale features and simultaneously preserve the detail information at each scale, which is helpful for the image fusion task.
\item A simple and efficient fusion network is proposed based on this framework, and excellent results can be achieved even only with a very simple fusion strategy, avoiding the complex problem of manually designing the fusion strategy.
\item Qualitative and quantitative experiments on multiple infrared and visible image fusion benchmarks demonstrate the superiority of the proposed method.
\end{itemize}
\par
The rest of this paper is organized as follows: Section \ref{sec-related} discusses related
work on image fusion. In Section \ref{sec-method}, we present the details of the proposed fusion
method. Section \ref{sec-experiment} shows the experimental results and compares them with some
state-of-the-art methods. Finally, we present the conclusions in Section \ref{sec-conclusion}.

\section{Related Works}\label{sec-related}
\subsection{Multi-scale transform based fusion methods}
Multi-scale decomposition is the most common feature extraction technique, such as Laplacian pyramid, wavelet transform~\cite{pajares2004wavelet}, complex wavelet transform~\cite{lewis2007pixel}, curvelet~\cite{nencini2007remote}, contourlet~\cite{do2002contourlets}, shearlet~\cite{easley2008sparse}, etc. Its basic step is first to obtain a multi-scale representation of the input image using a multi-scale transform. Then, the multi-scale representations of different images are fused according to specific fusion rules to obtain the fused multi-scale representation. Finally, the fused image is obtained by multi-scale inverse transform. In addition, there are some specific fusion rules, including choose-max~\cite{liu2015simultaneous}, weighted-average~\cite{yang2014visual}, optimization based method~\cite{ma2017infrared}, element-wise addition~\cite{ram2017deepfuse} and so on.\par
In general, most non-deep learning methods are time-consuming. The manually designed feature strategies cannot handle complex situations.


\subsection{Fusion methods based on deep learning}
Due to the powerful feature representation ability, deep learning is widely used in most computer vision tasks. A typical auto-encoder based fusion method is DenseFuse~\cite{li2018densefuse}. It uses dense blocks during encoding to extract features, where the output of each layer is adapted as the input to the next layer. Finally, the fusion image was reconstructed by the fusion strategy and the decoding network. In addition, AEFusion~\cite{li2023aefusion} captures long-range semantic information while extracting multi-scale features, combining with Axial-attention. A Transformer based auto-encoder presented, it uses self-supervised multi-task learning, called TransMEF~\cite{qu2022transmef}, which enables the network to deal with both local and global information at the same time. SEDRFuse~\cite{jian2020sedrfuse} uses a symmetric encoder-decoder and AUIF~\cite{zhao2021efficient} converts the two models into a basic encoder and a detail encoder. But these cannot avoid manually designing complex fusion strategies. \par

Therefore, other researchers focus on exploring the end-to-end CNN based image fusion networks, which rely on the complex network structure and the carefully designed loss function. A representative general fusion network is IFCNN~\cite{zhang2020ifcnn}. In addition, PSTL~\cite{xu2022infrared} deploys content branch and detail branch to extract characteristics and PIAFusion~\cite{tang2022piafusion} based on illumination-aware is proposed to take the illumination factor into account in the modeling process.\par

In recent years, many researchers have proposed Transformer-based infrared and visible image fusion algorithms. SwinFusion~\cite{ma2022swinfusion} proposes a general image fusion framework based on cross-domain long-range learning and Swin Transformer. SwinFuse~\cite{wang2022swinfuse} constructs a full attention feature encoding backbone to model long-range dependencies. This is a pure Transformer network. DATFuse~\cite{tang2023datfuse} uses a dual attention transformer which can extract important features and preserve global complementary information. Although the training of Transformer models requires high computer performance, more advanced Transformer-based algorithms may be proposed in the future due to the great scientific research potential of Transformers.\par

On the whole, most fusion algorithms based on deep learning do not change the spatial scale, and the feature richness is relatively insufficient. On the other hand, the method with multi-scale features only employs general pooling operations, which leads to the loss of information when extracting features. However, the lost information is beneficial to image generation and reconstruction, especially in the fusion task. This requires us to find other methods to combine with deep learning based methods.

\begin{figure*}[!htbp]
\begin{center}
\includegraphics[width=\textwidth]{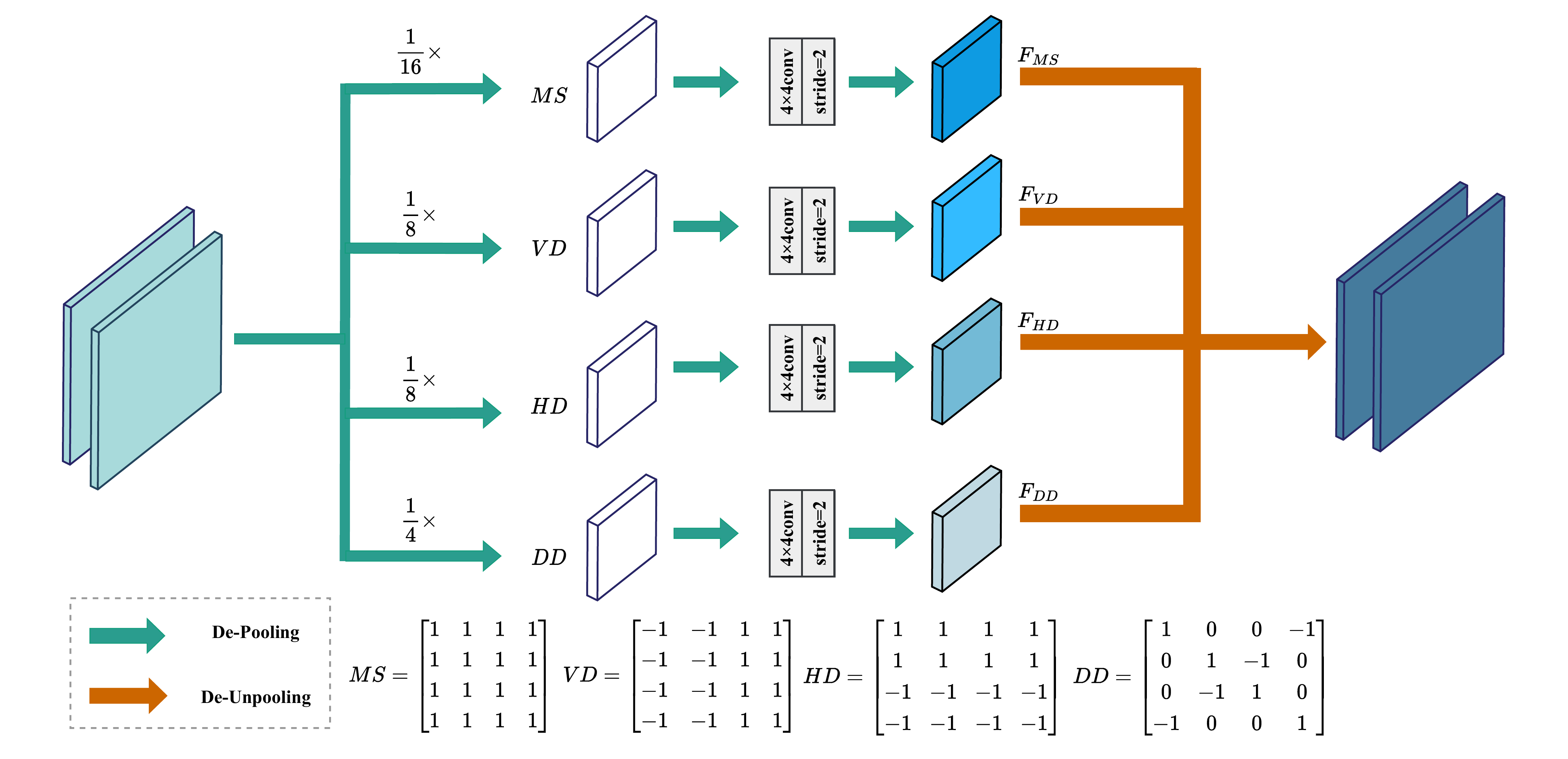}
\end{center}
\begin{center}
\vspace{-0.5cm}
\caption{The proposed module using decomposition pooling and unpooling. We employ $4\times4$ convolution kernel with stride 2 to extract multi-scale features $F_{MS}$ and detail features $\left( F_{VD}, F_{HD}, F_{DD}\right)$. Only the multi-scale component $MS$ passes to the encoding layer and the detail components $\left\{ VD, HD, DD\right\}$ are skipped to the corresponding decoding layer after feature fusion. In the decoder, the components are aggregated by the de-unpooling.
}\label{wave}
\end{center}
\vspace{-0.7cm}
\end{figure*}

\subsection{Combining multi-scale transform and deep learning}
When people pay too much attention to deep learning methods, they will ignore the ideas of non-deep learning methods. Therefore, more and more researchers have proposed a combination of non-deep learning and deep learning methods, which can use the excellent ideas of non-deep learning methods and the powerful framework of deep learning.

Chen~\cite{chen2023very} et al. propose a convolutional encoder–decoder image fusion network based on wavelet transform. They use max-pooling and up-sampling in the sub-network. The method proposed by Wang~\cite{wang2019multifocus} et al. uses convolutional neural networks in the discrete wavelet transform domain. They introduce the low-frequency information and high-frequency information of DWT into CNN-high and CNN-low networks to extract features, and similarly, they also use max-pooling operation in the network. Xu~\cite{xu2022multiscale} et al. proposed a multi-scale feature pyramid network based on activity level weight selection for infrared and visible image fusion. In this method, they use max-pooling and up-sampling convolution blocks to extract features and restore image details. Chao~\cite{chao2022medical} et al propose the discrete stationary wavelet transform (DSWT) to extract the high-frequency and low-frequency components of the image, and then they put them into radial basis function neural network (RBFNN) to extract features. Finally, the methods perform the inverse operation to reconstruct the image. \par

Although these methods can extract effective information in some scenarios, most of them use max-pooling to extract features, which leads to information loss and introduces noise. \par

The wavelet-based pooling work \cite{williams2018wavelet} is proposed as an alternative to traditional neighborhood pooling. However, they reduce the feature dimension by discarding the first-level components, while our method needs to utilize all the components. In LDWPooling~\cite{wang2021ldw}, although the proposed Learning Discrete Wavelet Pooling can be very good for image recognition and detection, it is not good enough for fine details processing, and the post-processing stage requires fine selection of key features and representative features, which is too complex.\par

In contrast, most pooling operations either cause information loss or extract part of the key features, which cannot adapt to complex multi-modal image fusion tasks. Our proposed de-pooling fully extracts multi-scale information while preserving texture details at each scale. In addition, de-unpooling can completely reconstruct the features without any post-processing operations and finally generate a fused image containing rich information.\par

\section{Proposed Fusion Method}\label{sec-method}
In this section, the proposed fusion method will be presented. Firstly, we give the details of decomposition pooling. Then, the model architecture is introduced. Finally, the details of training phase will be given.

\subsection{Decomposition Pooling}\label{depooling}
Compared with the common pooling operation, the decomposition pooling contains four kernels, $\left\{ MS, VD, HD, DD\right\}$, which represent four feature spaces respectively. As shown in Figure~\ref{wave}, the proposed de-pooling also contains four parts, the multi-scale component $MS$ captures global multi-scale information, while the detail components $VD$, $HD$ and $DD$ extract vertical, horizontal, and diagonal texture details.\par

Inspired by the Prewitt operator, our four feature spaces are formulated as following:
\begin{equation}
{\scriptsize MS=\begin{bmatrix}
  1&  1&  1&1\\
  1& 1 & 1 &1 \\
  1& 1 & 1 &1 \\
  1& 1 & 1 &1
\end{bmatrix}} \quad
{\scriptsize VD=\begin{bmatrix}
  -1&  -1&  1&1\\
  -1& -1 & 1 &1 \\
  -1& -1 & 1 &1 \\
  -1& -1 & 1 &1
\end{bmatrix}} 
\end{equation}

\begin{equation}
    {\scriptsize HD=\begin{bmatrix}
  1&  1&  1&1\\
  1& 1 & 1 &1 \\
  -1& -1 & -1 &-1 \\
  -1& -1 & -1 &-1
\end{bmatrix}}
{\scriptsize DD=\begin{bmatrix}
  1&  0&  0&-1\\
  0& 1 & -1 &0 \\
  0& -1 & 1 &0 \\
  -1& 0 & 0 &1
\end{bmatrix}} 
\end{equation}

\begin{figure}[!htbp]
\vspace{-0.1cm}
\includegraphics[width=0.75\textwidth]{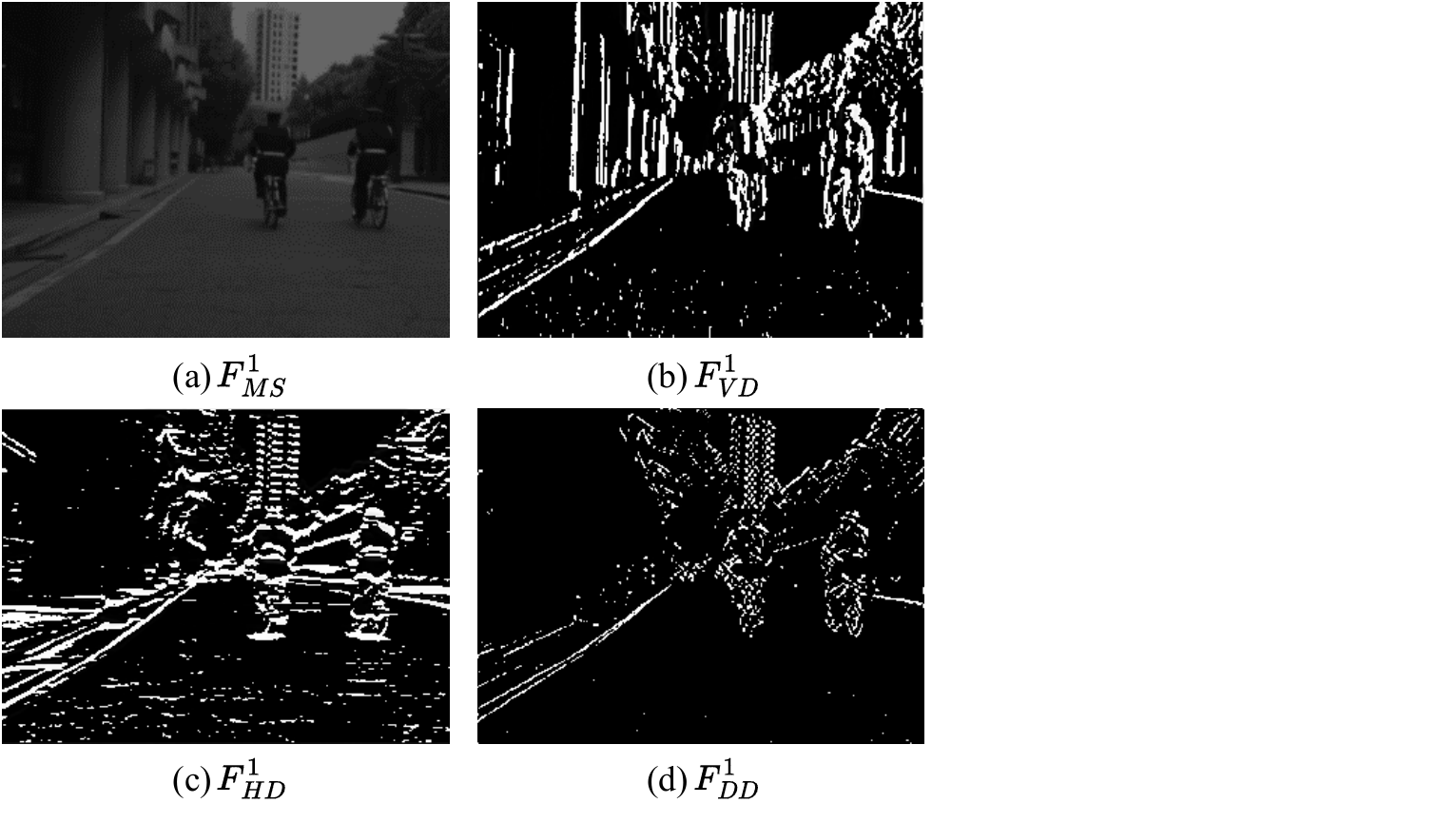}
\vspace{-0.8cm}
\caption{Visualization results of four feature maps after the first de-pooling. The global multi-scale information is obtained by multi-scale component $MS$, and the detail components $VD$, $HD$, $DD$ capture vertical, horizontal, and diagonal texture details.}\label{LL}
\end{figure}

In our proposed fusion network, we replace the common max-pooling layer with our de-pooling. The multi-scale components are passed to the encoding layer through convolutional blocks and decomposition pooling while de-unpooling utilizes the detail components of the source images for feature reconstruction. The four feature maps after the first de-pooling are shown in Figure~\ref{LL}. Specifically, the $MS$ component extracts multi-scale information, and the detail components $VD$, $HD$ and $DD$ extract texture details in three directions (vertical, horizontal and diagonal).\par

\begin{figure*}[!htbp]
\includegraphics[width=\textwidth]{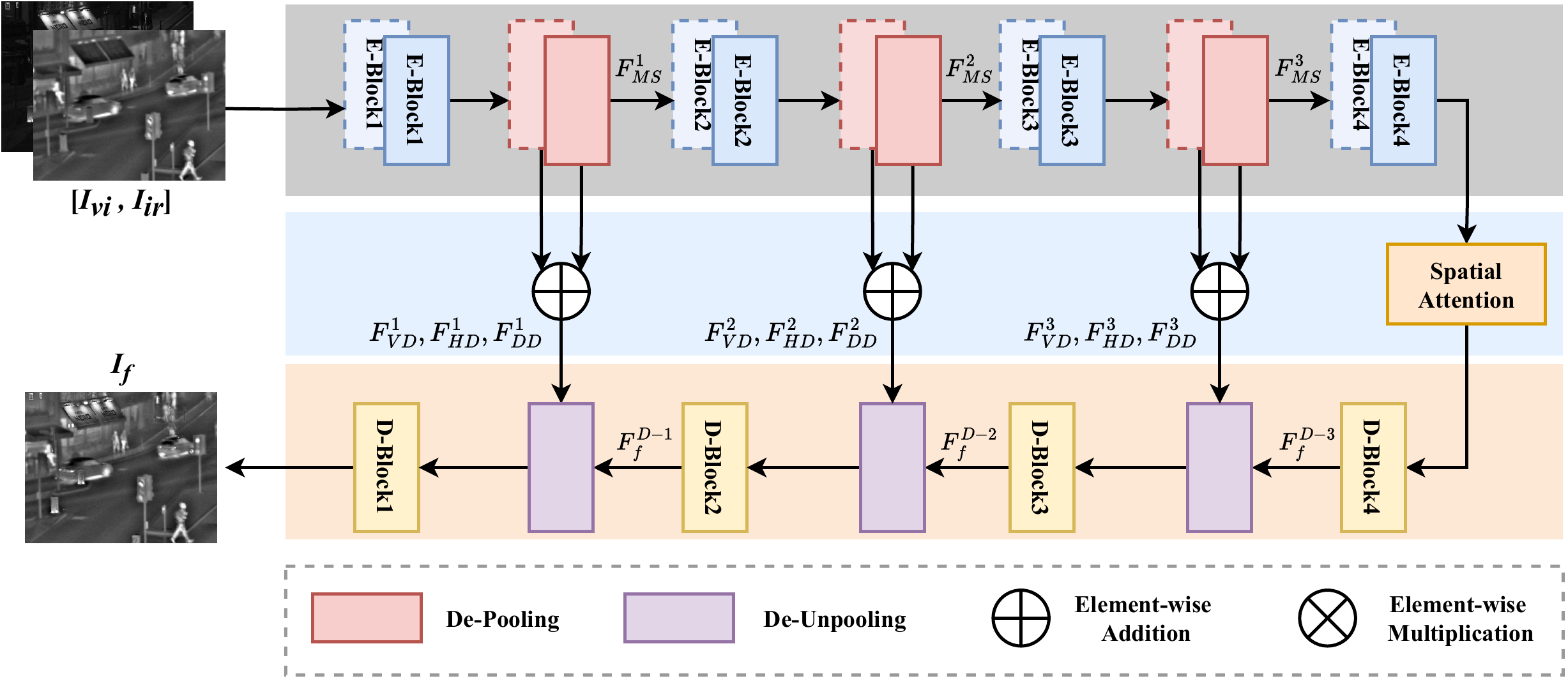}
\vspace{-0.5cm}
\caption{The architecture of the proposed method. A pair of encoder and decoder are at same scale. The encoder includes four convolutional blocks and three de-pooling, and the decoder consists of four convolutional blocks and three de-unpooling operations. De-unpooling is the inverse operation of de-pooling. The fusion layer uses spatial attention model (details in \ref{strategy}) and the detail information $\left\{ F_{VD}^{i}, F_{HD}^{i}, F_{DD}^{i}\right\}$ extracted by $i$th de-pooling employs addition strategy. $F_{MS}^{i}$ represents the multi-scale features extracted after the $i$th de-pooling. $F_{f}^{D-i}$ denotes the feature information reconstructed by $i$th de-unpooling together with the detail information.}\label{framework}
\vspace{-0.2cm}
\end{figure*}

Our de-unpooling has the property of minimal information loss and can reconstruct the images exactly. On the other hand, since max-pooling does not have its exact inverse, the networks proposed by Chen~\cite{chen2023very} et al. and Wang~\cite{wang2019multifocus} et al. cause greater information loss during the reconstruction process.\par


\subsection{Network Architecture} 
In order to fully extract the salient information and texture details from the multi-modal image and reconstruct the structure information of the source image, a novel fusion network architecture based on de-pooling is proposed. Our network architecture is composed of three parts: encoder, fusion layer, and decoder. The architecture is shown in Figure~\ref{framework}. The input infrared and visible images are denoted as $I_{vi}$ and $I_{vi}$. Note that the input images are pre-registered. The source images are processed by de-pooling to extract multi-scale information and simultaneously preserve detail information. In addition, the spatial attention model serves as our fusion strategy and each scale of the detail components corresponds to the addition. Finally, the extracted features are used to reconstruct the image through de-unpooling.\par

\subsubsection{Encoder}
The proposed encoder is shown in Figure~\ref{framework} (grey box). The solid and dashed lines in the Block represent the same structure and the same parameters. It has four convolutional blocks and three de-pooling operations. The proposed de-pooling can generate two pieces of information simultaneously: multi-scale structure information and detail information (contains vertical, horizontal, and diagonal texture details). Within the encoder, we employ de-pooling to extract multi-scale information while preserving texture details for later reconstruction, which can generate better images.\par

\begin{figure}[!htbp]
\vspace{-0.6cm}
\includegraphics[width=0.5\textwidth]{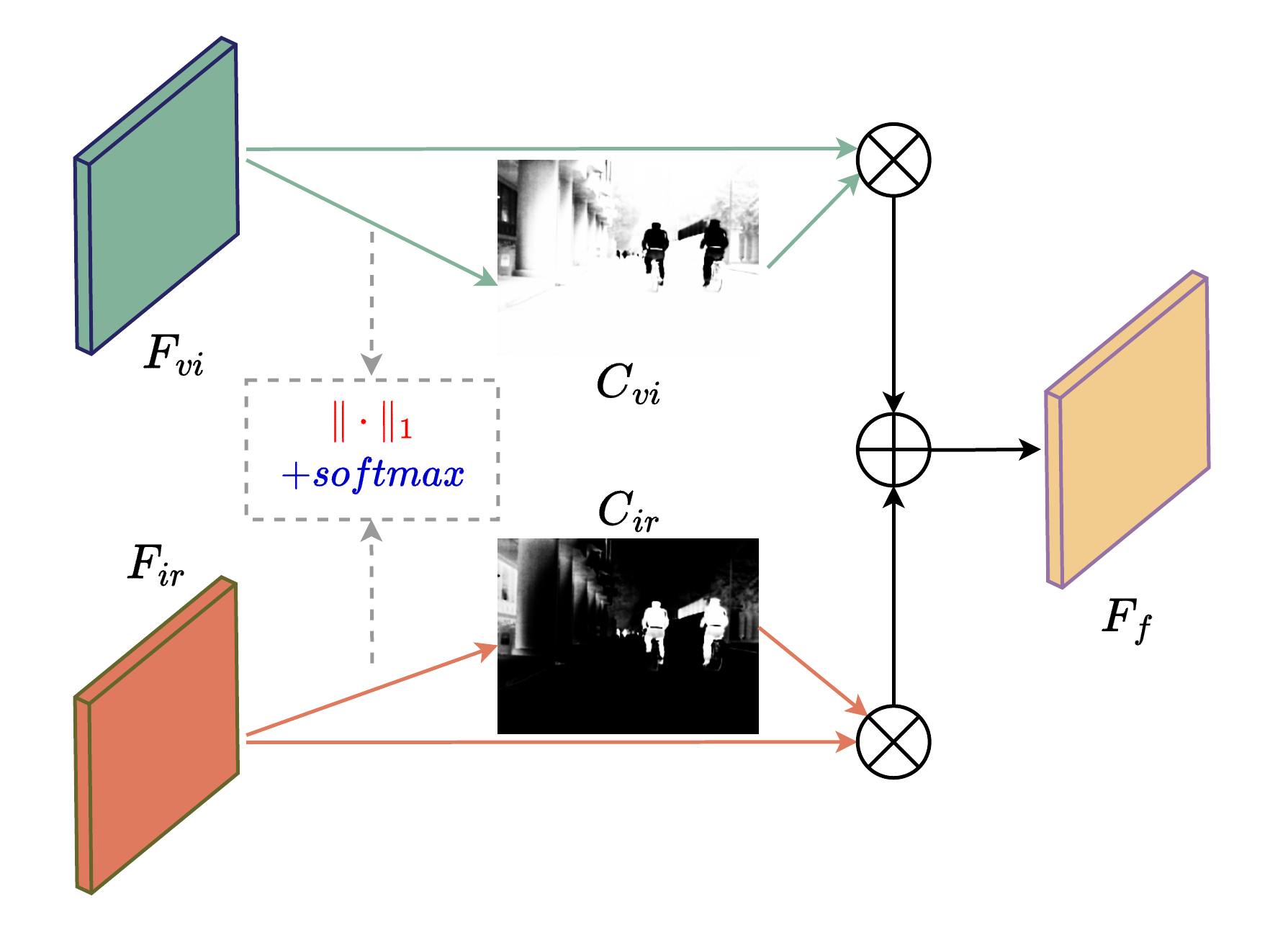}
\vspace{-0.8cm}
\caption{Procedure of spatial attention fusion strategy.}\label{l1}
\vspace{-0.2cm}
\end{figure}




\subsubsection{Fusion Strategy}\label{strategy}
\textbf{Multi-scale Features Fusion Strategy:} A well-designed fusion strategy is the key to the image fusion task. In this work, we introduce spatial attention model~\cite{li2020nestfuse,li2018infrared} into our method to fuse multi-scale deep features, in which the attention weights are calculated adaptively. The procedure for obtaining the spatial attention model is shown in Figure~\ref{l1}.\par

Given the multi-scale deep features $F_{ir} $ and $F_{vi} $ which are extracted by the last block of encoder (``E-Block4'') from infrared and visible images respectively, the weighting maps which are calculated
by $l_1$-normal and soft-max operator can be calculated by: 
\begin{equation}
C_{\omega}(x,y)=\frac{\Vert F_{\omega}(x,y) \Vert_1}{\sum\limits_{i\in \left\{ir, vi \right\}} \Vert F_{i}(x,y) \Vert_1}, \quad \omega \in \left\{ir, vi \right\}
\end{equation}
where $\Vert \cdot \Vert_1$ denotes $l_1$-normal. $(x,y)$ indicates the corresponding position in multi-scale deep features ($F_{ir}$ and $F_{vi}$) and weighting maps ($C_{ir}$ and $C_{vi}$ ).\par

Finally, the fused feature map $F_f$ is shown in the following equation:
\begin{equation}
F_f=\sum \limits_{i\in \left\{ir, vi \right\}} C_{i}(x,y) \times F_{i}(x,y)
\end{equation}

\textbf{Detail Features Fusion Strategy:} In addition, each scale corresponds to the detail information. Since the detail components are sparse, which are also very important for the generated image texture. Thus, in our framework, we are willing to preserve it as much as possible. The detail features fusion strategy is formulated as follows:

\begin{equation}
F_m^{i}=F_m^{ir}+F_m^{vi},\quad m \in \left\{VD, HD, DD \right\}
\end{equation}
Where $F_m^{ir}$ and $F_m^{vi}$ represent the texture details of infrared and visible images in vertical, horizontal, and diagonal directions, respectively. $F_m^{i}$ denote the final generated detail information through $i$th pooling layer in the three directions.

\par

\subsubsection{Decoder}
As shown in Figure~\ref{framework} (yellow box), the decoder reflects the encoder structure with four convolutional blocks and three de-unpooling. For better reconstruction features, we adopt summation for unpooling. Specifically, our de-unpooling performs the inverse operation of the processed detail features and multi-scale features output by de-pooling from the corresponding scales. Therefore, the decoder with de-unpooling is able to completely reconstruct the image features.\par

\begin{table}[!htbp]
\renewcommand{\arraystretch}{1.2}
\vspace{-0.1cm}
\centering
\belowrulesep=0pt
\caption{Network architecture of our method. \textbf{Input} and \textbf{Output} denote the number of channels in the corresponding feature maps.}\label{tab1}
\scalebox{0.9}{
\begin{tabular}{c|cccccc}
\midrule 
& Blocks &
Layers &
Kernel & 
Input &
Output &
Activation\\
\hline 
\multirow{10}{*}{Encoder} 
&\multirow{3}{*}{E-Block1}& Layer1 & $1\times1$ & 1 & 1 &  ReLU\\
&\multirow{3}{*}{}& Layer2 & $3\times3$ & 1 & 64 &  ReLU\\
&\multirow{3}{*}{}& Layer3 & $3\times3$ & 64 & 64 &  ReLU\\ \cline{2-7}

&\multirow{2}{*}{E-Block2}& Layer1 & $3\times3$ & 64 & 128 &  ReLU\\
&\multirow{2}{*}{}& Layer2 & $3\times3$ & 128 & 128 &  ReLU\\\cline{2-7}

&\multirow{4}{*}{E-Block3}& Layer1 & $3\times3$ & 128 & 256 &  ReLU\\
&\multirow{4}{*}{}& Layer2 & $3\times3$ & 256 & 256 &  ReLU\\
&\multirow{4}{*}{}& Layer3 & $3\times3$ & 256 & 256 &  ReLU\\
&\multirow{4}{*}{}& Layer4 & $3\times3$ & 256 & 256 &  ReLU\\\cline{2-7}

&\multirow{1}{*}{E-Block4}& Layer1 & $3\times3$ & 256 & 512 &  -\\

\hline 
\multirow{9}{*}{Decoder} 

&\multirow{1}{*}{D-Block4}& Layer1 & $3\times3$ & 512 & 256 &  -\\\cline{2-7}

&\multirow{4}{*}{D-Block3}& Layer1 & $3\times3$ & 256 & 256 &  ReLU\\
&\multirow{4}{*}{}& Layer2 & $3\times3$ & 256 & 256 &  ReLU\\
&\multirow{4}{*}{}& Layer3 & $3\times3$ & 256 & 256 &  ReLU\\
&\multirow{4}{*}{}& Layer4 & $3\times3$ & 256 & 128 &  ReLU\\\cline{2-7}

&\multirow{2}{*}{D-Block2}& Layer1 & $3\times3$ & 128 & 128 &  ReLU\\
&\multirow{2}{*}{}& Layer2 & $3\times3$ & 128 & 64 &  ReLU\\\cline{2-7}

&\multirow{2}{*}{D-Block1}& Layer1 & $3\times3$ & 64 & 64 &  ReLU\\
&\multirow{2}{*}{}& Layer2 & $3\times3$ & 64 & 1 &  ReLU\\ 
\hline 
\end{tabular}}
\vspace{-0.6cm}
\end{table}

\subsection{Training Phase}
\begin{figure}[!htbp]
\begin{center}
\includegraphics[width=0.5\textwidth]{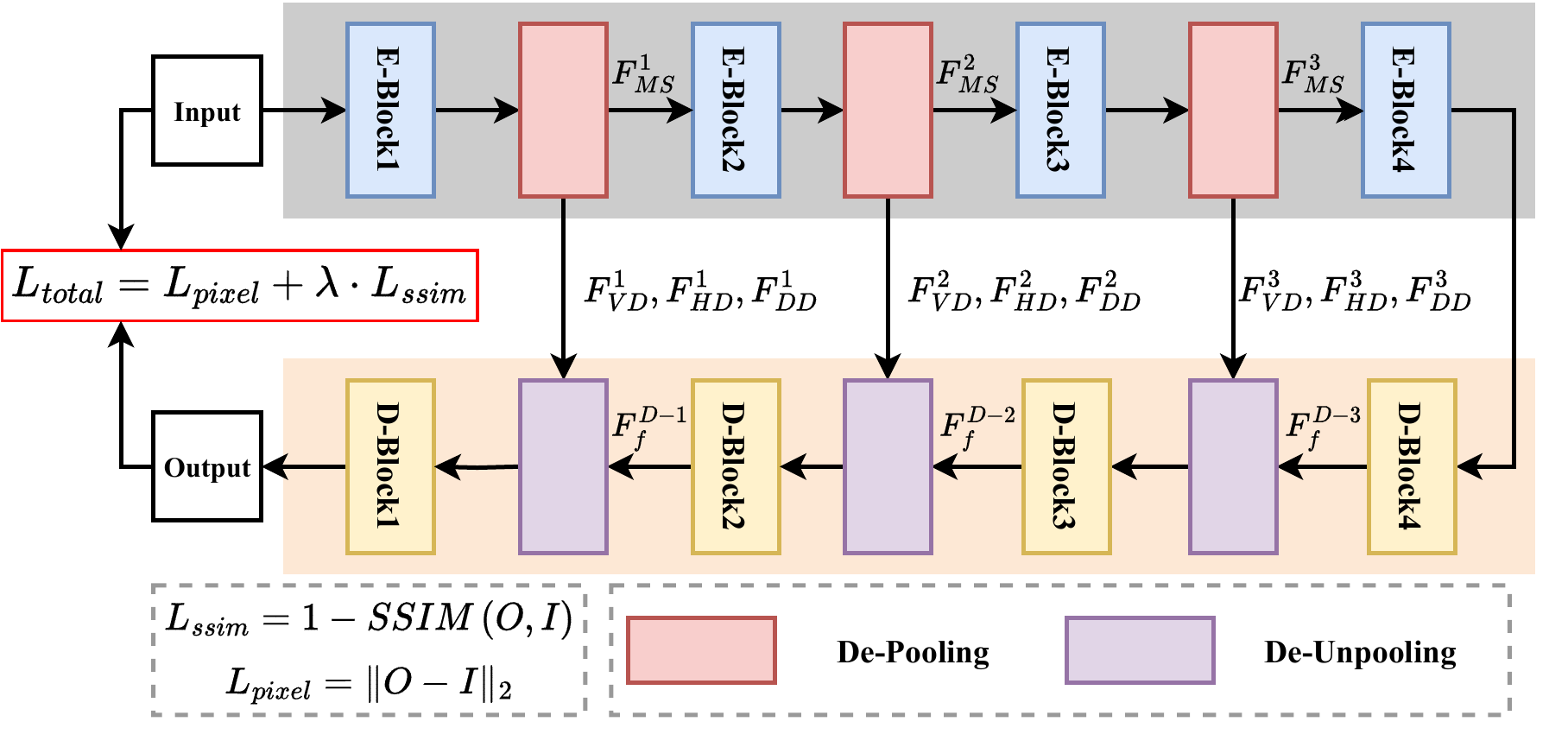}
\end{center}
\begin{center}
\caption{The framework of training process. In this process, we train an auto-encoder network without fusion strategy.}\label{auto}
\end{center}
\vspace{-0.5cm}
\end{figure}

Our training phase is based on the auto-encoder training strategy, and the fusion strategy is discarded in this phase. With this strategy, the encoder is able to extract multi-scale structural information and texture details from input images, the decoder can well reconstruct the input from the extracted information. The training framework is shown in Figure~\ref{auto}, and the network architecture is shown in Table ~\ref{tab1}.\par
In order to reconstruct the input image more precisely, the loss function $L_{total}$ is defined as follows:
\begin{equation}
L_{total}=L_{pixel} + \lambda L_{ssim}
\label{loss}
\end{equation}
where $L_{pixel}$ and $L_{ssim}$ indicate the pixel loss and structure similarity (SSIM) loss, respectively. $\lambda$ denotes the tradeoff value between $L_{pixel}$ and $L_{ssim}$.\par

$L_{pixel}$ is calculated by the following equation:
\begin{equation}
L_{pixel}= \Vert O - I \Vert_F^2
\end{equation}

where $O$ and $I$ indicate the output and input images, respectively. $\Vert \cdot \Vert_F^2$ is the $l_2$ norm. This loss function will make sure that the reconstructed image is more similar to the input image at the pixel level. \par
The SSIM loss $L_{ssim}$ is formulated as follows:

\begin{equation}
L_{ssim}=1-SSIM\left( O,I \right)
\end{equation}
where $SSIM(\cdot)$ denotes the structural similarity between the generated images and source images~\cite{wang2004image}.

\begin{figure*}[!htbp]
\includegraphics[width=\textwidth]{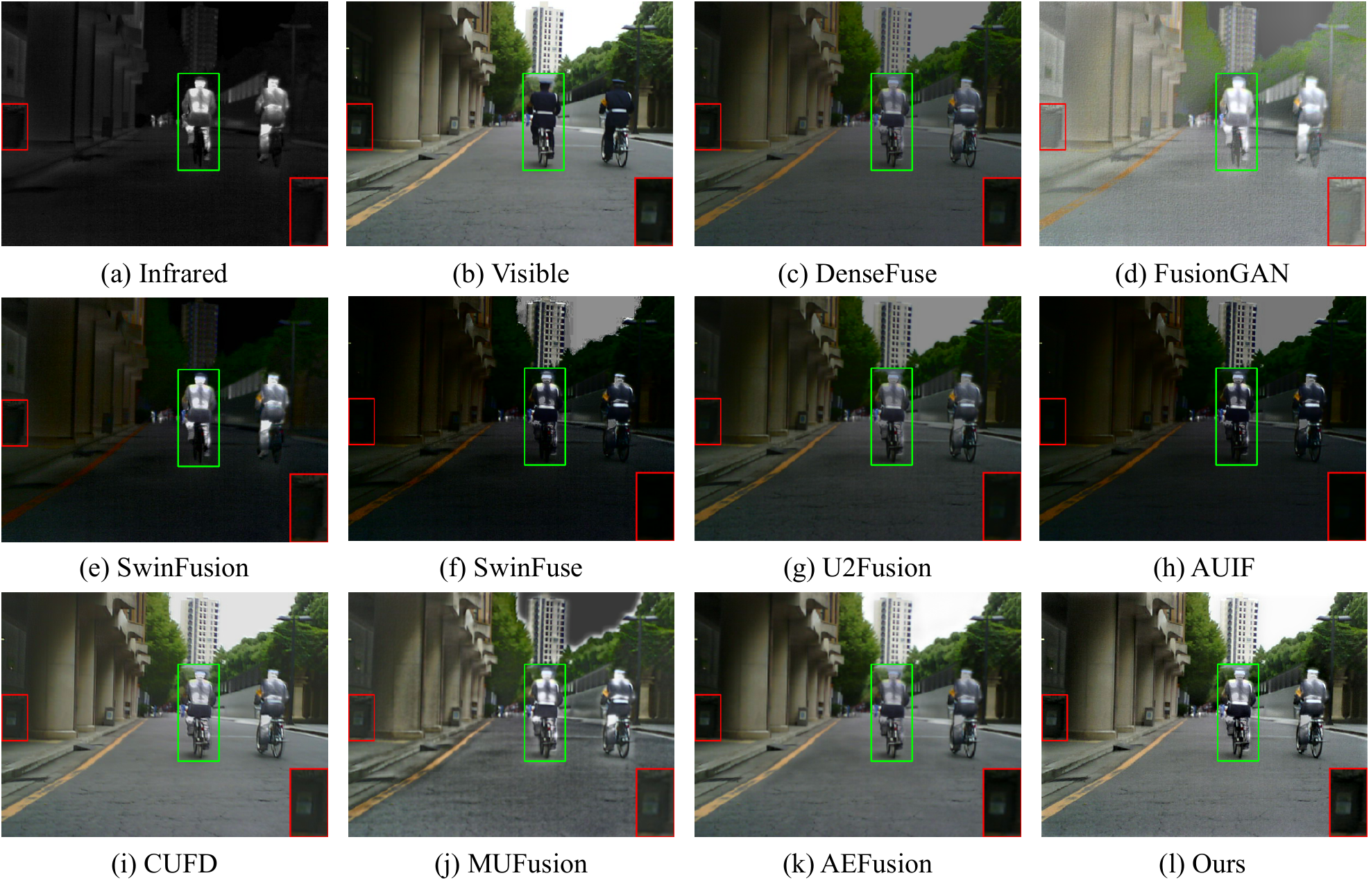}
\vspace{-0.8cm}
\caption{Qualitative comparison of our method with nine state-of-the-art methods on the MSRS dataset.}\label{compare_msrs}
\vspace{-0.4cm}
\end{figure*}

\section{Experimental Results and Analysis}\label{sec-experiment}
\subsection{Experimental Settings} 
In the training phase, we randomly choose 40000 images from MS-COCO~\cite{lin2014microsoft} dataset to train our auto-encoder network, all of them are resized to $256 \times 256$. The learning rate is set as $1 \times 10^{-4}$. The batch size and epochs are set to 4 and 4, respectively. The hyperparameter $\lambda$ in Equation~\ref{loss} is set as 100. All the involved experiments are conducted on an NVIDIA RTX 3090Ti GPU and Intel Core i7-10700 CPU. To fuse the RGB images, firstly, the visible images are converted to the YCbCr color space. Then, the Y channel of the visible images and the infrared images are fused by the proposed method. Finally, the fused image is converted back to the RGB color space by concatenating the Cb and Cr channels of the visible images.\par

To comprehensively evaluate the proposed algorithm, we perform qualitative and quantitative experiments on the MSRS dataset~\cite{tang2022piafusion} with all the 361 image pairs, the LLVIP dataset~\cite{jia2021llvip} with randomly selected 389 image pairs and the TNO dataset~\cite{toet2017tno} with randomly selected 16 image pairs.\par 

To evaluate the fusion performance, we choose nine typical and state-of-the-art fusion methods, including DenseFuse~\cite{li2018densefuse}, FusionGAN~\cite{ma2019fusiongan}, SwinFusion~\cite{ma2022swinfusion}, SwinFuse~\cite{wang2022swinfuse}, U2Fusion~\cite{xu2020u2fusion}, AUIF~\cite{zhao2021efficient}, CUFD~\cite{xu2022cufd}, MUFusion~\cite{cheng2023mufusion} and AEFusion~\cite{li2023aefusion}. The implementations of these approaches are publicly available.\par

Five quality metrics are utilized for quantitative comparison between our fusion method and other existing fusion methods, including standard deviation (SD), visual information fidelity (VIF), average gradient (AG), the sum of correlations of differences (SCD) and entropy (EN). 
SD reflects the visual effect of the fused image. VIF measures the information fidelity of the fused image. AG quantifies the gradient information of the fused image and represents its detail and texture. SCD reflects the level of correlation between the information transmitted to the fused image and corresponding source images. EN is used to represent the image detail retention. 
Moreover, a fusion algorithm with larger SD, VIF, AG, SCD and EN indicates better fusion performance.

\subsection{Comparative experiments}
In order to fully demonstrate the superiority of our approach, we compare our method with other nine SOTA methods on MSRS, LLVIP and TNO datasets.

\subsubsection{Fusion Results on MSRS Dataset}
Qualitative experiments performed on the MSRS dataset are shown in Figure~\ref{compare_msrs}. As shown in the red highlighted regions, SwinFusion, SwinFuse, U2Fusion and AUIF can barely preserve the detail information of the visible image. In addition, FusionGAN suffers from severe spectral contamination, which degrades the visual effect of the fused image. Although DenseFuse preserves part of the texture details, it weakens salient features, resulting in an overall darker effect. In the green box, we can find that MUFusion blurs the edges of targets and introduces a lot of artifacts and noise into results in some regions, causing visual conflicts. In contrast to our results, although CUFD and AEFusion integrate the texture information of visible images with salient target information in infrared images to some extent, they produce abundant artifacts that blur edges and details, making the overall effect is inferior to ours. In general, only our method successfully maintains the structure intensity and preserves the textures, thanking to our wavelet pooling and unpooling.\par

\begin{figure*}[htbp]
\centering
\includegraphics[width=0.95\textwidth]{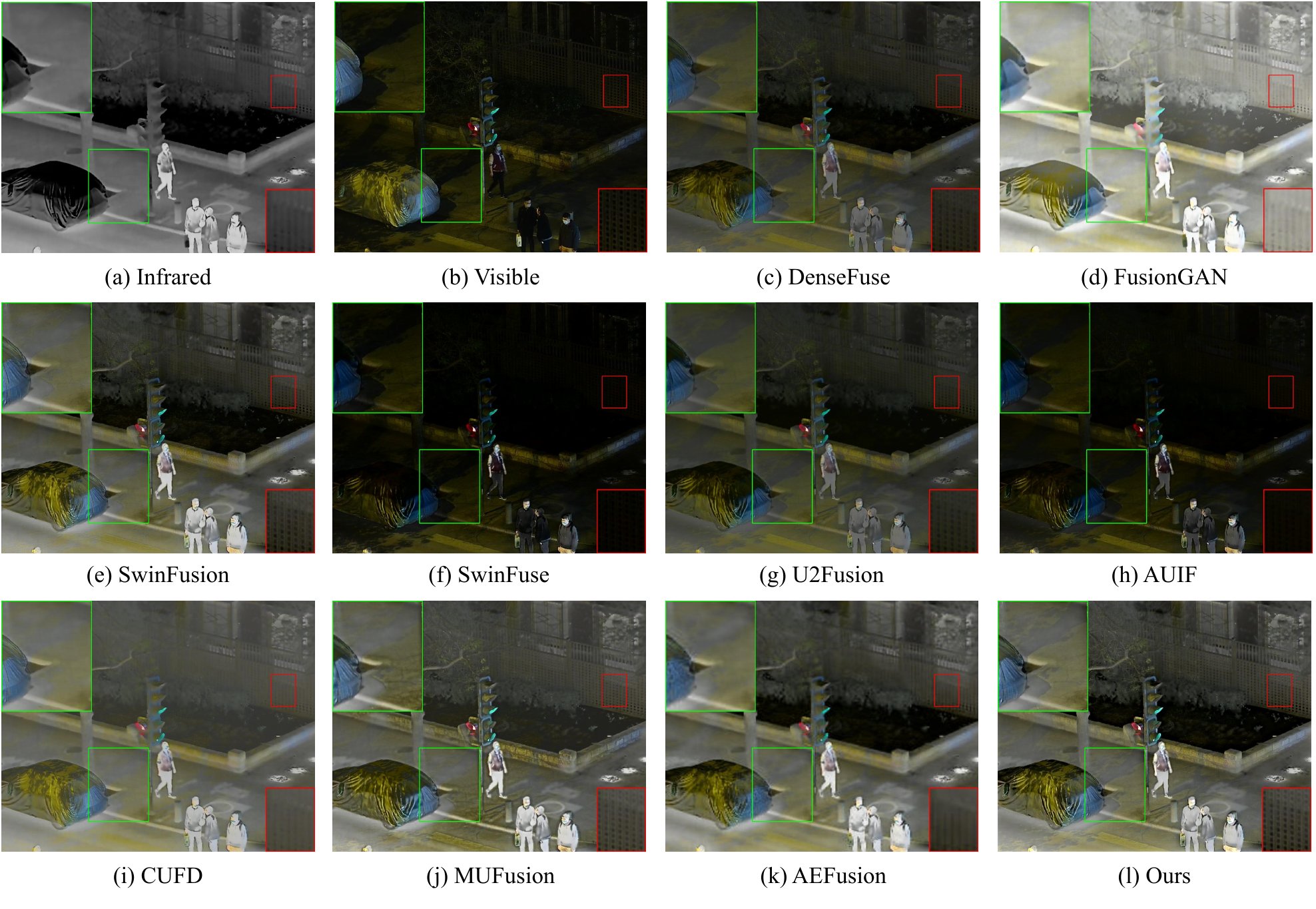}
\vspace{-0.5cm}
\caption{Qualitative comparison of our method with nine state-of-the-art methods on the LLVIP dataset.}\label{compare_llvip}
\vspace{-0.3cm}
\end{figure*}

\begin{table}[htbp]
\vspace{-0.3cm}
\renewcommand{\arraystretch}{1.7}
\centering
\caption{Quantitative results on 361 image pairs from the MSRS dataset. (\textbf{Bold}: Best, \textcolor{red}{Red}: Second Best, \textcolor{blue}{Blue}: Third best)}
\scalebox{1}{
\begin{tabular}{c|ccccc}
\hline 
\centering
Methods& SD  & VIF & AG &SCD &EN  \\
\hline 
DenseFuse & 7.4237   & 0.6999  & 2.0873  &  1.2489 & 5.9340  \\
\hline 
FusionGAN	& 7.1758     & \textcolor{blue}{0.8692}   &  \textcolor{blue}{3.1193}  & 0.3129  &5.9937  \\

\hline 
SwinFusion	& 6.0518 & \textcolor{red}{0.9382} & 1.7343 & 0.5631 &5.2846   \\

\hline 
SwinFuse & 4.9246  & 0.4102   & 1.9673    &1.0129  &4.4521     \\

\hline 
U2Fusion	 & 6.8217    & 0.5863   &  2.0694  & \textcolor{blue}{1.2955} &5.5515   \\

\hline 
AUIF	 &  5.2622 & 0.3981 &1.8238 &1.0639 &4.6460   \\

\hline 
CUFD	 &\textcolor{blue}{7.6384} & 0.6488 &2.9003 &1.2379 & \textcolor{blue}{6.0652}     \\

\hline 
MUFusion & 6.9233 & 0.6086   &\textcolor{red}{3.1474} & 1.2548 &5.9682   \\

\hline 
AEFusion	 &\textcolor{red}{8.2104} &0.8548 &2.6968 &\textcolor{red}{1.4564} &\textcolor{red}{6.5374}    \\

\hline 

Ours  & \textbf{8.4779} &\textbf{0.9508} &\textbf{4.0498} &\textbf{1.7159}&\textbf{6.7697} \\
\hline  

\end{tabular}
\label{table_msrs}}
\end{table}

The quantitative results of seven metrics on the MSRS dataset are presented in Table~\ref{table_msrs}. The subjective comparison in Figure~\ref{compare_msrs} presents the distinctive superiority of our method in SD and VIF metrics, indicating that our results have high contrast, maximization of the information fidelity and satisfying visual effects, which is consistent with the human visual system. 
For the AG metric, the first ranking of our method illustrates that our fusion results contain more edge information.
From the results, we can see that our framework ranks first in SCD and EN metrics, which demonstrates that our fusion results retain realistic and valid information correlated with the source images. \par

\begin{figure*}[!htbp]
\centering
\includegraphics[width=0.9\textwidth]{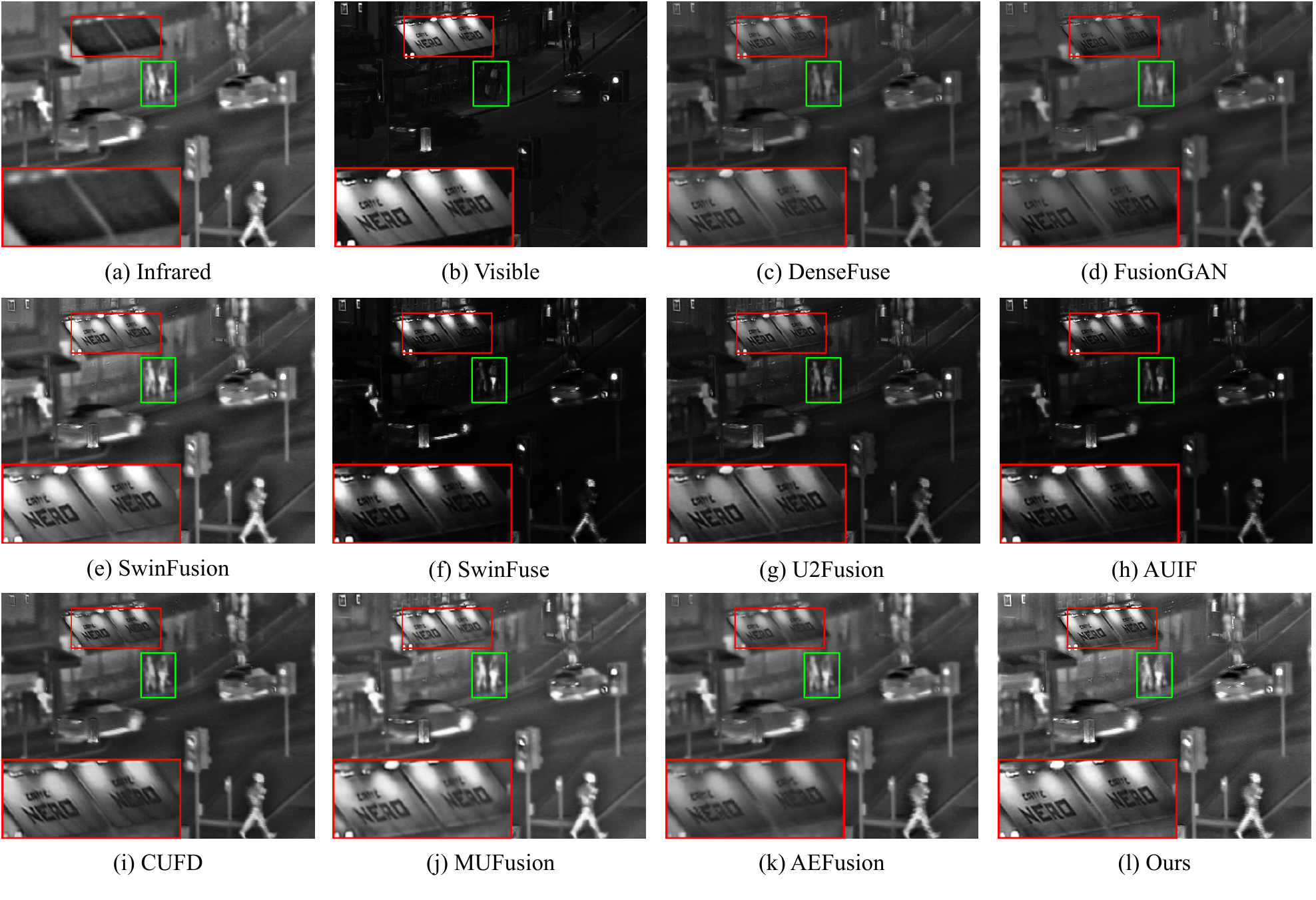}
\vspace{-0.5cm}
\caption{Qualitative comparison of our method with nine state-of-the-art methods on the TNO dataset.}\label{compare_tno}
\vspace{-0.3cm}
\end{figure*}
\subsubsection{Fusion Results on LLVIP Dataset}
The visualization results of different methods on the LLVIP dataset are shown in Figure~\ref{compare_llvip}. The selected nighttime image can demonstrate the superiority of our approach. As can be seen from the red box, SwinFuse and AUIF fail to retain detail information of the fence. Although FusionGAN, CUFD and AEFusion maintain the infrared salient features, the background regions are all affected by varying degrees of spectral contamination, resulting in blurring detail information. In addition, DenseFuse and U2Fusion weaken the infrared features, making the overall images darker and affecting human visual perception. Only our method, SwinFusion and MUFusion effectively preserve the texture details of the fence. As observed from the green box, SwinFusion and MUFusion do not deal well with the details of branch shadows. They introduce artifacts and almost blend in with the ground without a sharp distinction. The same goes for other methods except our method. Our method does a better job in terms of significant target maintenance and details preservation than other methods.\par

\begin{table}[htbp]
\renewcommand{\arraystretch}{1.7}
\centering
\caption{Quantitative results on 389 image pairs from the LLVIP dataset. (\textbf{Bold}: Best, \textcolor{red}{Red}: Second Best, \textcolor{blue}{Blue}: Third best)
}
\scalebox{1}{
\begin{tabular}{c|ccccc}
\hline
Methods& SD  & VIF & AG &SCD &EN  \\
\hline 
DenseFuse &9.2963 &0.7503 &2.6714 &1.2109 &6.8287    \\
\hline 
FusionGAN	&\textbf{10.0823} &\textbf{1.0263} &2.1706 &0.4276 &7.1741   \\

\hline 
SwinFusion	&9.6297 &\textcolor{red}{0.9807} &\textcolor{red}{4.3851} &\textcolor{red}{1.5678} &\textcolor{red}{7.3466 }  \\

\hline 
SwinFuse &7.5469 &0.6290 &2.9580 &1.1840 &6.0825      \\

\hline 
U2Fusion	&9.4256 &0.7212 &2.3685 &1.3114 &6.7588      \\

\hline 
AUIF	 &7.5433 &0.5877 &2.8790 &1.1667 &6.1555    \\

\hline 
CUFD	 &9.1701 &0.7187 &2.5198 &1.0360 &6.8448      \\

\hline 
MUFusion &8.7452 &0.7875 &\textcolor{blue}{3.5412} &1.0975 &6.9242   \\

\hline 
AEFusion	 &\textcolor{blue}{9.8400} &0.6302 &2.0397 &\textcolor{blue}{1.3526} &\textcolor{blue}{7.2764}   \\

\hline 


Ours  &\textcolor{red}{9.9393} &\textcolor{blue}{0.9725} &\textbf{4.7475} &\textbf{1.6155} &\textbf{7.4398}  \\
\hline
\end{tabular}
\label{table_llvip}}

\end{table}

The comparative subjective results of different methods on the LLVIP dataset are shown in Table~\ref{table_llvip}. Our method achieves the best results on 3 of these 5 metrics. The highest performance on 
AG proves that our fusion results are able to preserve rich gradient information. Meanwhile, the best results in SCD and EN metrics mean that our fusion results are highly consistent with the source images and can generate more realistic images. 
The SD metric of our method ranks second because FusionGAN introduces more noise and artifacts into the fused image. Moreover, we can also observe that FusionGAN in the VIF metric is optimal, but the texture information is lost in visualization, resulting in blurred details and image distortion. \par

\subsubsection{Fusion Results on TNO Dataset}
The qualitative comparisons of different algorithms on the TNO dataset are presented in Figure~\ref{compare_tno}. As shown in the green boxes, DenseFuse, SwinFuse, U2Fusion and AUIF weaken the salient target. In addition, it can be seen from the red box that the overall images of SwinFuse and AUIF are darker than others and are unable to preserve the texture details of the visible images. For FusionGAN and AEFusion, they blur the sharp edges of visible images. In addition, CUFD and MUFusion introduce many artifacts into the fused images, resulting in worse visual perception. Only our method and SwinFusion are able to maintain structure intensities while preserving texture details.\par

\begin{table}[htbp]
\centering
\caption{Quantitative results on 16 image pairs from the TNO dataset. (\textbf{Bold}: Best, \textcolor{red}{Red}: Second Best, \textcolor{blue}{Blue}: Third best)
}
\renewcommand{\arraystretch}{1.7}
\scalebox{1}{
\begin{tabular}{c|ccccc}
\hline 
\centering
Methods& SD  & VIF & AG &SCD &EN  \\
\hline 
DenseFuse &9.2203 &0.7349 &3.8804 &1.6300 &6.8256  \\
\hline 
FusionGAN	&8.1234 &0.6197 &2.8120 &1.1911 &6.4629   \\

\hline 
SwinFusion	&\textcolor{blue}{9.5370} &\textbf{0.8907} &5.3161 &\textcolor{blue}{1.6855} &7.0270  \\

\hline 
SwinFuse &9.2633 &\textcolor{blue}{0.7982} &\textcolor{red}{5.5986} &\textcolor{red}{1.6882} &6.9484   \\

\hline 
U2Fusion &9.3869 &0.7200 &5.4456 &1.6406 &6.9395    \\

\hline 
AUIF	 &9.2805 &0.7482 &5.2820 &\textbf{1.7537} &7.0402  \\

\hline 
CUFD	 &9.4136 &\textcolor{red}{0.8781} &4.5178 &1.4135 &\textcolor{blue}{7.0743}  \\

\hline 
MUFusion &\textcolor{red}{9.5379} &0.7851 &\textcolor{blue}{5.5756} &1.5338 &\textbf{7.3032} \\

\hline 
AEFusion	 &9.4655 &0.7803 &3.4114 &1.5744 &7.0716  \\
\hline


Ours  &\textbf{9.6036} &0.7786 &\textbf{6.0003} &1.5964 &\textcolor{red}{7.1583}  \\
\hline
\end{tabular}}
\label{table_tno}
\vspace{-0.5cm}
\end{table}

We use quantitative metrics to measure the performance of different methods on the TNO dataset , which is shown in Table~\ref{table_tno}. The best results in AG and SD mean that our method captures abundant texture details and simultaneously achieves the best visual perceptual performance. 
Our method trails MUFusion by a narrow margin in the EN metric, but our method does not introduce artifacts and noise into the fusion results from visualizing results and also contains a lot of meaningful information transmitted from the source image to the fused image.
However, for the VIF and SCD metrics, our method does not perform as well as on the other two datasets. This is justified since the source images in TNO dataset contain less detail information. Compared with other datasets (with RGB images), the visible images in TNO mainly have better infrared salient features, while texture details are not obvious. However, de-pooling prefers to preserve the texture details and the intensity of source images. Thus, in TNO dataset, our proposed method achieves comparable fusion performance.\par

In conclusion, both qualitative and quantitative results demonstrate that our method achieves better/comparable fusion performance. Moreover, our method has obvious advantages in preserving texture details and maintaining structure information, and can achieve a pleasing visual perception.\par

\subsection{Ablation Studies}
\begin{table}[!htbp]
\renewcommand{\arraystretch}{1.8}
\centering
\caption{Quantitative comparison of $2\times2$ convolution kernels and $4\times4$ convolution kernels in MSRS, LLVIP and TNO datasets.(\textbf{Bold}: Best)}
\label{table1}
\scalebox{1}{
\begin{tabular}{c|c|c|c|c|c|c|c}
\hline

\multicolumn{2}{c|}{\multirow{2}{*}{Metrics}}  & \multicolumn{2}{c}{MSRS Dataset}&\multicolumn{2}{c}{LLVIP Dataset} &\multicolumn{2}{c}{TNO Dataset}\\
\cline{3-8}

\multicolumn{2}{c|}{} &$2\times2$ & $4\times4$&$2\times2$ & $4\times4$ &$2\times2$ &$4\times4$ \\
\hline

\multicolumn{2}{c|}{SD} & 8.4648 &	\textbf{8.4779}	
&\textbf{9.9500} &	9.9393	
&9.5748 &	\textbf{9.6036} 	\\ 
\hline

\multicolumn{2}{c|}{VIF} & \textbf{0.9570} &	0.9508	
&0.9682 	&\textbf{0.9725}	
&\textbf{0.7887} &	0.7786	\\ 
\hline

\multicolumn{2}{c|}{AG} &3.9988 &	\textbf{4.0498}	
&4.7143 	&\textbf{4.7475} 	
&5.9975 &	\textbf{6.0003}	\\ 
\hline

\multicolumn{2}{c|}{SCD} &\textbf{1.7358} &	1.7159	
&1.6134 &	\textbf{1.6155}	
&\textbf{1.6394} &	1.5964	\\ 
\hline

\multicolumn{2}{c|}{EN} &6.7486 & \textbf{6.7697}	
&7.4315 &	\textbf{7.4398}	
&7.1394 &	\textbf{7.1583}	\\ 
\hline



\end{tabular}}
\label{2compare4}
\end{table}
\subsubsection{Analysis of the difference kernel of de-pooling} \label{fields}
In our fusion framework, the proposed de-pooling uses $4\times4$ convolution kernels instead of $2\times2$ convolution kernels~\cite{yoo2019photorealistic}. The fundamental difference in the size of the convolution kernel is the different receptive fields. The receptive field reflects the perception range of the current input feature map of the convolution kernel. If the range is small, the received information is one-sided and local. As the receptive field increases, more global information can be obtained, which is more conducive to the judgment of the current situation. Therefore, a large convolution kernel leads to a larger receptive field. However, blindly increasing the convolution kernel will bring the problem of increasing the amount of calculation. Therefore, it is necessary to choose the appropriate convolution kernel in deep learning.\par 
As shown in Table~\ref{2compare4}, the metrics of our $4\times4$ convolution kernel are better than the $2\times2$ convolution kernel in three metrics on the MSRS dataset and TNO dataset, and in four metrics on the LLVIP dataset. This fully demonstrates that the de-pooling we designed can better contain rich information and texture details, while achieving the best visual quality.

\begin{figure}[!htbp]
\includegraphics[width=0.75\textwidth]{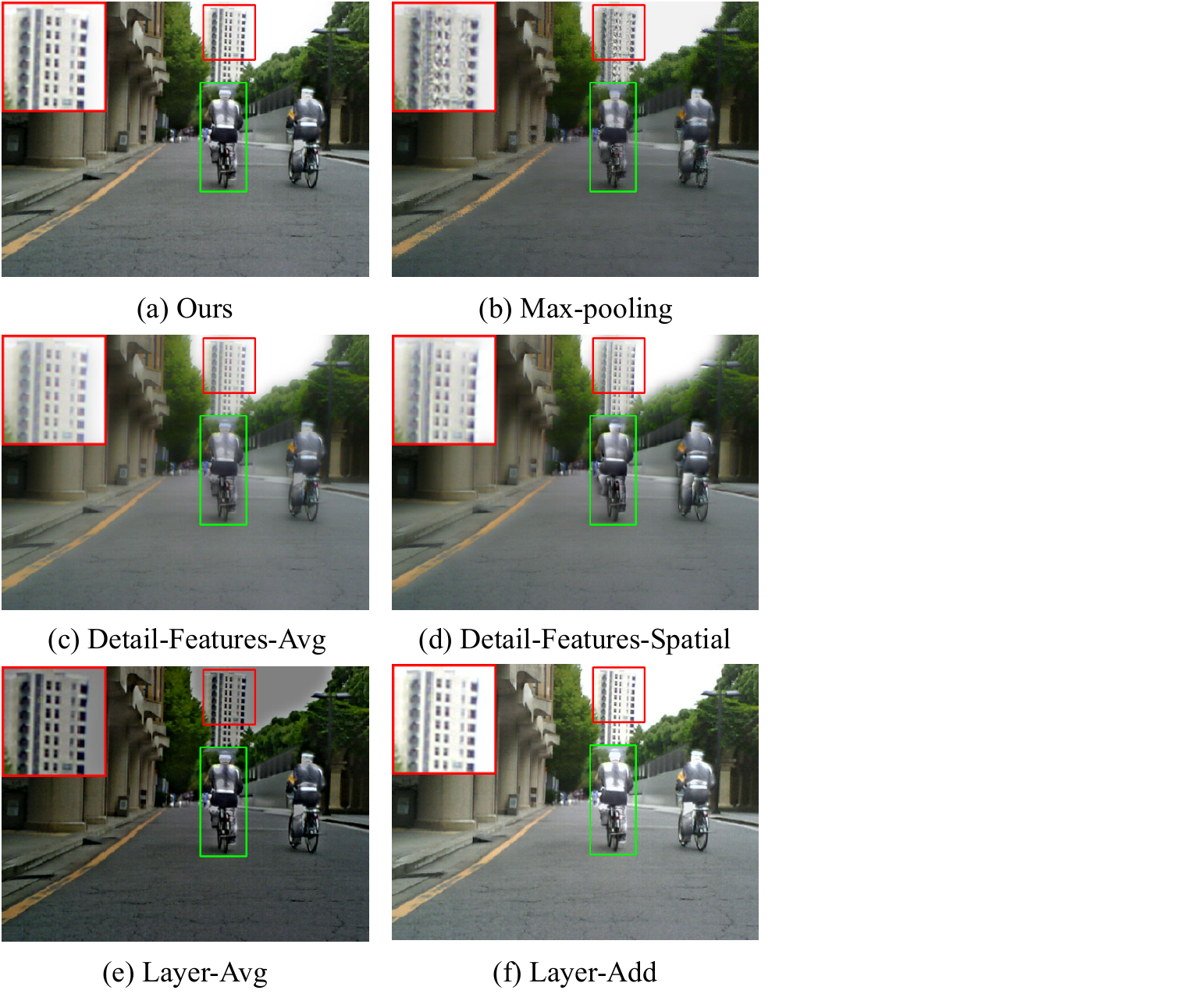}
\vspace{-0.8cm}
\caption{Visualized results of ablation studies. From (a) to (f): fused results of our method, fused results of using max-pooling, detail features with different directions using average strategy, detail features with different directions using spatial attention fusion strategy, fusion layer using average strategy and fusion layer using addition strategy.}\label{xiaorong}
\vspace{-0.1cm}
\end{figure}

\subsubsection{Max-pooling vs De-pooling}
To demonstrate that our model does benefit from decomposition pooling, we compare fusion results using max-pooling. As shown in Figure~\ref{xiaorong} (b), although max-pooling can extract the salient features, it leads to degradation in fine details such as buildings (red box). In addition, the max-pooling introduces a lot of artifacts and noise, which is also because the max-pooling does not have its exact inverse. The proposed de-pooling can not only successfully preserves fine details and maintain structure information, but also has the characteristics of minimum information loss, allowing the network to completely reconstruct the signal. The quantitative results in Table~\ref{table_ablation} show that our method outperforms max-pooling in all metrics

\begin{table}[htbp]
\renewcommand{\arraystretch}{1.9}
\centering
\caption{The average values of the five objective metrics obtained with different ablation studies on MSRS dataset. (\textbf{Bold}: Best, \textcolor{red}{Red}: Second Best, \textcolor{blue}{Blue}: Third best)
}
\scalebox{0.95}{
\renewcommand{\arraystretch}{1.9}
\begin{tabular}{c|ccccc}
\hline 
\centering
Strategies & SD  & VIF & AG &SCD &EN  \\
\hline 
Ours &\textcolor{red}{8.4779} &\textbf{0.9508} &\textbf{4.0498} &\textcolor{red}{1.7159}& \textcolor{red}{6.7697}  \\

\hline 
Max-Pooling	&7.8781 &0.7087 &2.6391 &1.3769 &6.3626   \\

\hline 
Detail-Features-Avg	&8.3962 &0.6736 &2.1711 &1.3599 &6.5538  \\

\hline 
Detail-Features-Spatial &\textcolor{blue}{8.4381} &0.5920 &2.3306 &1.2897 &\textcolor{blue}{6.5724}   \\

\hline 
Layer-Avg &7.6953 &\textcolor{red}{0.8894} &\textcolor{red}{4.0208} &\textcolor{blue}{1.4381} &6.3852    \\
 
\hline 
Layer-Add &\textbf{8.6109} &\textcolor{blue}{0.8464} &\textcolor{blue}{3.8939} &\textbf{1.8974} &\textbf{6.8248}  \\

\hline 

\end{tabular}
\label{table_ablation}}
\vspace{-0.5cm}
\end{table}

\subsubsection{Analysis of detail features fusion strategy}
In our method, the detail information $\left( VD, HD, DD\right)$ extracted by de-pooling is added correspondingly, and then the de-unpooling is guided by this information to reconstruct features. To prove that the addition operation can better preserve edge information, we compare this strategy to the average strategy and spatial attention fusion strategy in this experiment. As shown in Figure~\ref{xiaorong} (c) and (d), average strategy and spatial attention fusion strategy for detail features will weaken the structure information and blur the edge of background regions in the red box. Moreover, these two kinds of strategies introduce a large number of artifacts and achieve a bad visual effect. It can be seen that the detail features with addition strategy can maintain structure intensities and strengthen edge information. The quantitative results are shown in Table~\ref{table_ablation} and our method achieves the best results.

\subsubsection{Analysis of multi-scale features fusion strategy}
For the last layer of the encoder, we use the simple spatial attention fusion strategy to fuse multi-scale features. The spatial attention fusion strategy is utilized to fuse multi-scale deep features. In this ablation study, the average strategy and the addition strategy are conducted. As shown in Figure~\ref{xiaorong} (e) and (f), the average strategy weakens the background brightness of the visible image and leads to an overall darker image. However, the addition strategy will excessively enhance the salient features of the infrared image, resulting in visual conflict. It can be observed that the spatial attention fusion strategy can better integrate multi-scale features and achieve satisfying visual senses.\par
The average values of the five metrics are shown in Table~\ref{table_ablation}. Our method has better performances than using average strategy. Compared with the addition strategy, only VIF and AG metrics of our method are better than it. However, it can be observed from the visualization results that the addition strategy will enhance the salient features too much and form over-exposed images. Although this strategy improves the metrics from introducing too much information, our method can achieve better visual effects, which is why we need to choose multiple metrics from different dimensions to evaluate the quality of the generated images.

\section{Conclusion}\label{sec-conclusion}
In this work, a novel decomposition pooling based fusion network for infrared and visible image fusion is proposed, termed as DePF. Our network has three main parts: encoder, fusion layer and decoder. The proposed de-pooling does not follow the traditional method based on wavelet pooling, but expands the receptive field and extracts more global deep information. Moreover, the structure of the feature extraction kernel is changed based on Prewitt operation. Within the de-pooling, it can extract multi-scale information while preserving more detail information. It is also worth emphasizing that the exact recovery of the de-unpooling enables our model to successfully maintain structure intensities, preserve fine details and reconstruct images. In addition, our fusion network adopts the spatial attention fusion strategy, which is simple and efficient. To verify the effectiveness of the proposed method, we perform our results on three publicly available datasets. Our method achieves excellent performance in terms of visualization effects and quantitative evaluation than nine state-of-the-art methods. Ablation experiments also demonstrate the effectiveness of different components of the proposed method.




%


\ifCLASSOPTIONcaptionsoff
  \newpage
\fi



%
\bibliographystyle{IEEEtran}
\bibliography{main}

%




\end{document}